\documentclass[preprint,12pt]{elsarticle}

\usepackage{amssymb}
\usepackage{graphicx}
\usepackage{url}
\def\*#1{\mathbf{#1}}
\usepackage{amsmath}

\usepackage{xcolor}
\usepackage{algorithm}
\usepackage{algpseudocode}
\usepackage{booktabs}
\usepackage[colorlinks=true, linkcolor=blue, citecolor=blue, urlcolor=blue]{hyperref}

\journal{Medical Image Analysis}

\begin{document}

\begin{frontmatter}



\title{VIDS-Seg: Towards Reliable Uncertainty Quantification in Pediatric Cardiac Ultrasound Segmentation}


\author[bas]{Paul Fischer} 
\ead{paul.fischer@unibas.ch}
\author[bas]{Ece Özkan Elsen}

\affiliation[bas]{organization={Department of Biomedical Engineering, University of Basel},
            city={Basel},
            country={Switzerland}}

\begin{abstract}
Reliable clinical deployment of machine learning requires models that know when they are likely to fail, particularly for subgroups underrepresented in training data. A common case is pediatric care, where models trained on adult cohorts can silently under-perform on children with no indication that something has gone wrong. As retraining with labeled pediatric data is often infeasible, detecting such failures at inference time is a critical clinical need. Building on the VIDS (Variational Inference under Distribution Shifts) framework, we introduce \textbf{VIDS-Seg}, which applies amortized variational inference over a lightweight prediction head to make this adaptive, OOD-aware prior tractable for dense image segmentation. We evaluate VIDS-Seg on left ventricular segmentation in echocardiography, a setting where pediatric anatomy differs systematically from the adult population most segmentation models are trained on, training on an adult cohort (EchoNet-Dynamic) and evaluating zero-shot on a pediatric cohort (EchoNet-Pediatric). Across all age strata, VIDS-Seg matches competitive baselines in segmentation accuracy while producing substantially higher spatial correspondence between predicted uncertainty and segmentation error, an advantage that persists even after applying temperature scaling to all baselines. Downstream, it yields more accurate and stable ejection fraction estimates and more reliable detection of cardiac malfunction in the infant subgroup. Our results indicate that OOD-aware uncertainty quantification can serve as a practical safety layer for deployed segmentation models, enabling detection of silent failures in underrepresented subgroups without retraining or additional labeled data.
\end{abstract}




\end{frontmatter}



\section{Introduction}
\label{sec:introduction}
Machine learning is rapidly becoming a practical tool in clinical medicine. The number of AI-enabled medical devices cleared by the U.S. Food and Drug Administration has grown substantially over the past decade \citep{FDA-AIML-Database-2023}, spanning applications from radiology and pathology to cardiology. This clinical adoption brings new demands for reliability: a model that performs well on the population it was trained on may fail systematically when deployed on a different one. A particularly consequential example of such a distributional mismatch arises when models trained predominantly on adult cohorts are applied to pediatric patients \citep{chatterjee2025children}. Children differ substantially from adults in anatomy, physiology, and image acquisition characteristics \citep{lopez2024guidelines}. A natural response would be to train and evaluate models specifically on pediatric data. In practice, however, this is rarely sufficient. Large, richly annotated pediatric datasets are scarce relative to their adult counterparts, and even where pediatric data exist, children are not a homogeneous population. Anatomy and physiology change rapidly from infancy through adolescence, so a model trained broadly on ``pediatric'' data can itself exhibit the same silent under-performance on specific pediatric subgroups, such as infants, that motivates this work in the first place. The challenge is therefore not specific to the adult–child boundary but reflects a more general problem: detecting when a model is unreliable for any subgroup underrepresented in its training distribution. Deploying an adult-trained model in a pediatric setting without awareness of this mismatch can produce systematic, clinically relevant errors. Left undetected, such errors may lead to incorrect diagnoses, biased treatment decisions, and unequal care for an already vulnerable population \citep{joseph2025multi,muralidharan2023recommendations}. This effectively establishes a two-tier standard of care, raising both clinical safety and robustness concerns.
 
Out-of-distribution (OOD) detection and uncertainty quantification (UQ) have attracted considerable attention in the machine learning community as potential tools for identifying exactly this kind of failure \citep{yang2024generalized,hong2024out,zou2023review}. OOD detection methods aim to flag inputs that lie far from the training distribution, while UQ methods aim to provide calibrated confidence estimates that track the model's actual error rate. Together they offer the prospect of models that ``know what they don't know''. In clinical imaging, uncertainty estimates are increasingly being studied not merely as a technical property but for their direct clinical utility such as flagging uncertain predictions for expert review, identifying cases that require additional data acquisition, or triggering escalation to specialist care \citep{han_uncertainty_2021}. Alongside this, there is growing recognition that proper evaluation of uncertainty is essential. Properties such as calibration, coverage, and spatial correspondence between uncertainty and error are now actively studied \citep{lambert2022improving}.
 
A critical but underappreciated limitation of most UQ methods, however, is that their desirable properties, calibration, coverage, and appropriate uncertainty magnitude, are typically established and evaluated on in-distribution test data. There is no general mechanism by which standard UQ methods should produce \emph{higher} uncertainty for inputs that lie outside the training distribution \citep{ovadia2019can}. This is especially problematic when the OOD group is also the group with the worst predictive performance: the model fails silently, producing confident but incorrect predictions. In practice, an adult-trained cardiac segmentation model deployed in a pediatric clinic may produce plausible-looking segmentations for most children while systematically failing for the youngest and most anatomically atypical patients, without any indication in the model's output that something has gone wrong. Even where labeled pediatric data can be obtained, annotation is costly and may not be available for every relevant subgroup at the point of deployment \citep{wang2024artificial,gonzalez2022distance}. In many clinical settings, the realistic question is therefore not how to eliminate every possible performance gap through retraining, but how to reliably \emph{detect} such gaps at inference time so that appropriate human oversight can be triggered. Detecting such silent, systematic failures is not merely a technical curiosity but rather a prerequisite for the robust and safe deployment of machine learning in diverse clinical populations \citep{lekadir2025future}.
 
The recently introduced VIDS framework \citep{slavutsky2025quantifying} offers a promising approach to this problem. By introducing an adaptive prior that conditions on test covariates, VIDS is explicitly designed to increase uncertainty for inputs that deviate from the training distribution, addressing the silent-failure problem at its root. However, VIDS was introduced and evaluated only for classification and regression tasks. Segmentation, often used in medical image analysis for analyses such as volume estimation, disease staging, and treatment planning, was not considered. Adapting VIDS to dense pixel-level prediction is non-trivial. Applying amortized variational inference over the full parameter space of a modern segmentation network is computationally intractable, and the spatial structure of both inputs and outputs demands new design choices in the embedding and energy-function components of the framework.
 
In this work we address these gaps. Our main contributions are:

\begin{itemize}
    \item \textbf{VIDS-Seg}: We extend the VIDS framework to dense segmentation by applying amortized variational inference over a lightweight prediction head. This decomposition makes the variational problem tractable while preserving the OOD-awareness of the adaptive prior, and requires no retraining of the core segmentation network.
    \item \textbf{Empirical OOD identification and uncertainty evaluation}: We systematically evaluate uncertainty quality across age-stratified pediatric subgroups, identifying infants as the OOD population and demonstrating that VIDS-Seg produces significantly better spatial correspondence between uncertainty and prediction error compared to other popular baselines.
    \item \textbf{Downstream clinical impact}: We show that the higher-quality uncertainty from VIDS-Seg translates into more accurate and stable ejection fraction estimates for the OOD subgroup, illustrating that uncertainty quality has direct consequences for clinical decision-making derived from model predictions.
    \item \textbf{Positioning relative to post-hoc calibration}: We show that pure post-hoc calibration cannot substitute for an OOD-aware prior under unknown subgroup shift and that VIDS-Seg composes with rather than competes against them.
\end{itemize}

\section{Related Work}
\label{sec:related_work}
\subsection{Probabilistic and Uncertainty-Aware Segmentation}
Probabilistic segmentation methods aim to produce uncertainty estimates alongside predictions. PHiSeg \citep{shen_phiseg_2019} introduced a hierarchical latent variable model to capture multi-scale aleatoric uncertainty arising from annotation ambiguity. \cite{lambert2024trustworthy} provide a comprehensive review of uncertainty quantification in deep learning for medical image analysis, noting that ensemble methods and Monte Carlo dropout remain the most widely deployed approaches in practice. A shared limitation of all these methods is that their uncertainty properties are typically validated on in-distribution data, without explicit treatment of covariate shift.

\subsection{Out-of-Distribution Detection in Medical Segmentation}
Several works have studied the problem of detecting OOD inputs specifically in the context of segmentation. \cite{lambert2022improving} evaluated a range of uncertainty frameworks for OOD detection in multiple sclerosis lesion segmentation, finding that predictive uncertainty from standard binary segmentation models often fails to detect OOD inputs reliably, while multi-label approaches show improvement. \cite{gonzalez2021self} explored self-supervised OOD detection for cardiac segmentation in Cine Magnetic Resonance Images, demonstrating that self-supervised signals can help identify distribution shifts without explicit OOD supervision. While these methods detect or adapt to distribution shift, none of them provides a principled probabilistic mechanism to condition uncertainty on the relationship between training and test distributions, which is the core contribution of our approach. VIDS \citep{slavutsky2025quantifying} takes a fundamentally different approach by introducing an adaptive prior over network parameters that is conditioned on both training and test covariates, explicitly modeling the dependence of uncertainty on distributional novelty.

\subsection{Relation to Post-hoc Calibration}
\label{sec:posthoc}

VIDS-Seg targets a different failure mode than post-hoc calibration methods, and the two are complementary rather than competing.

Temperature scaling and conformal prediction adjust an \emph{existing} uncertainty signal: a single scalar temperature, or a single conformal score threshold, is estimated on a calibration set and applied uniformly at test time. Two properties follow. First, per-pixel temperature scaling is a monotone transformation of the predicted confidence and therefore leaves the \emph{spatial ordering} of entropy within an image largely intact. It can correct the magnitude of uncertainty but not its location. Second, the coverage guarantees of split conformal prediction require exchangeability between the calibration and test sets. This is precisely the assumption violated under covariate shift. Class- or group-conditional variants restore guarantees only when subgroup membership is known at test time, which is exactly the information unavailable in the deployment scenario we consider: the model is not told that the patient is an infant. Conformalizing dense segmentation introduces further difficulty, as prediction sets must be defined over spatial structures rather than scalars \citep{mossina2025conformal}.

Our contribution is therefore situated upstream of calibration: we aim to improve the \emph{quality} of the base uncertainty estimate, in particular its spatial correspondence with error under shift, which any subsequent calibration step inherits. We make this concrete in Section~\ref{sec:uncertainty}, where we show that temperature scaling fitted on in-distribution (adult) data leaves the relative ordering of methods unchanged, and that VIDS-Seg remains the strongest base estimator with and without post-hoc adjustment.

\section{Methods}
\label{sec:methods}
 \begin{figure}[t]
    \centering
    \includegraphics[width=0.95\textwidth]{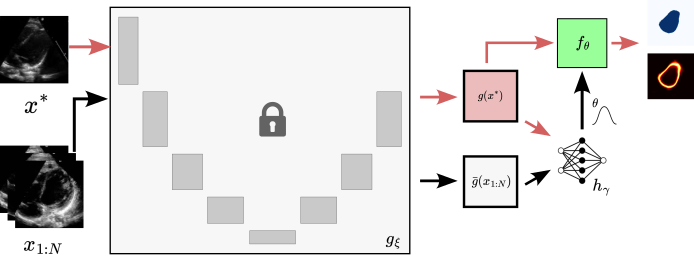}
    \caption{
        \textbf{VIDS-Seg overview.}
        The U-Net backbone (gray) is frozen.
        For each environment, training images are embedded and spatially pooled to form a context summary (mean and std across images).
        Each test image is also embedded and pooled.
        The inference network $h_\gamma$ takes the concatenated context and test embedding and outputs the parameters of the variational distribution $q_\phi(\theta)$ over the prediction-head parameters.
        The predictive entropy map serves as the pixel-level uncertainty estimate.
    }
    \label{fig:architecture}
\end{figure}
We build on VIDS \citep{slavutsky2025quantifying}, a Bayesian approach for uncertainty quantification under covariate shift, and extend it to the task of dense image segmentation. Section~\ref{sec:vids_background} summarizes the key ideas of VIDS that are necessary to understand our extension. Sections~\ref{sec:segmentation_extension}--\ref{sec:inference_procedure} describe VIDS-Seg.

\subsection{VIDS: Variational Inference under Distribution Shifts}
\label{sec:vids_background}
 
Let $\mathcal{D} = \{(x_i, y_i)\}_{i=1}^{N}$ denote the training set, where $x_i \in \mathcal{X}$ are input covariates and $y_i \in \mathcal{Y}$ are labels. Let $\theta \in \mathbb{R}^{d_\theta}$ denote the parameters of a predictive network $f_\theta$, and let $x^* \in \mathcal{X}$ be a new test covariate drawn from a potentially shifted distribution.

In a standard Bayesian neural network, the posterior predictive distribution is
\begin{equation}
    p(y^* \mid x^*, x_{1:N}, y_{1:N}) = \int p(y^* \mid x^*, \theta)\, p(\theta \mid x_{1:N}, y_{1:N})\, d\theta,
    \label{eq:classical_posterior_predictive}
\end{equation}
where $p(\theta \mid x_{1:N}, y_{1:N})$ is the posterior over parameters. Because $x^*$ does not enter the posterior, this formulation cannot raise uncertainty when $x^*$ lies far from the training distribution.

VIDS addresses this by replacing the fixed prior $p(\theta)$ with an \emph{adaptive prior} $p(\theta \mid x_{1:N}, x^*)$ conditioned on both training covariates and the test point, defined via an energy function:
\begin{equation}
    p(\theta \mid x_{1:N}, x^*) \propto \exp\!\left(\int \textstyle\sum_{i=1}^{N} \log p(y \mid x_i, \theta) + \log p(y \mid x^*, \theta)\, dy\right).
    \label{eq:adaptive_prior}
\end{equation}
This causes the posterior over $\theta$, and therefore the predictive uncertainty, to widen whenever $x^*$ is novel relative to $x_{1:N}$.
Since the resulting posterior is intractable, VIDS fits an amortized variational distribution $q_\phi(\theta; x^*)$, where the parameters $\phi = (\mu, \sigma^2)$ are produced by an \emph{inference network} $h_\gamma$:
\begin{equation}
    (\mu, \log\sigma) = h_\gamma\!\left(\bar{g}(x_{1:n}),\, \hat{g}(x^*)\right),
    \label{eq:inference_network}
\end{equation}
with $\hat{g} : \mathcal{X} \to \mathbb{R}^k$ a pre-trained embedding network, $\bar{g}(x_{1:n}) = \frac{1}{n}\sum_{i=1}^{n}\hat{g}(x_i)$ a permutation-invariant summary of the context set $\{x_i\}_{i=1}^n$, and $h_\gamma$ a fully-connected network with learnable weights $\gamma$.
The variational parameters are optimized by maximizing the Evidence Lower Bound (ELBO):
\begin{equation}
    \mathcal{L}(\phi;\, x^*, \mathcal{D}) = \mathbb{E}_{q_\phi}\!\left[\log p(y_{1:N} \mid x_{1:N}, \theta)\right] - \mathrm{KL}\!\left(q_\phi(\theta;\,x^*) \;\|\; p(\theta \mid x_{1:N}, x^*)\right).
    \label{eq:elbo}
\end{equation}
Since shifted test covariates are unavailable at training time, VIDS constructs $L$ \emph{synthetic environments} $\{e^{(\ell)}\}_{\ell=1}^{L}$ by drawing small bootstrap subsamples $(\mathcal{D}^{(\ell)}_\mathrm{tr}, \mathcal{D}^{(\ell)}_\mathrm{te})$ from $\mathcal{D}$, each of size $n$ and $m$ respectively.
Because low-probability subsamples deviate from the empirical distribution of $\mathcal{D}$, they simulate plausible covariate shifts.
The inference network is optimized over all environments using a cross-environment objective with a variance penalty \citep{krueger2021out}:
\begin{equation}
    \mathcal{L} = \frac{1}{L}\sum_{\ell=1}^{L} \mathcal{L}^{(\ell)} - \tau\, \mathrm{Var}\!\left(\mathcal{L}^{(1)}, \ldots, \mathcal{L}^{(L)}\right),
    \label{eq:cross_env_objective}
\end{equation}
where $\mathcal{L}^{(\ell)}$ is the ELBO on environment $\ell$, and $\tau > 0$ is the variance-penalty weight.
At test time, $S$ samples $\theta^{(s)} \sim q_\phi$ are drawn via the reparametrization trick ($\theta = \mu + \sigma \odot \varepsilon$, $\varepsilon \sim \mathcal{N}(0, I)$) and used to form the posterior predictive distribution.

\subsection{VIDS-Seg}
\label{sec:segmentation_extension}
 
The original VIDS framework was designed for classification and regression tasks, where the prediction network is a compact fully-connected model.
Applying amortized variational inference naively to a full segmentation network such as a U-Net is infeasible: a standard U-Net has on the order of $10^7$--$10^8$ parameters \citep{tran2021tmd}, making the variational posterior over $\theta$ intractably high-dimensional and the inference network $h_\gamma$ prohibitively large.
 
\paragraph{Variational inference on the prediction head only}
We observe that the vast majority of a U-Net's parameters are responsible for building a rich, spatially structured feature representation: the encoding and decoding pathway.
The final prediction (i.e., the assignment of a class to each pixel) is performed by a lightweight $1\!\times\!1$ convolution that maps the feature map to class logits.
We therefore decompose the network into two components with very different roles, and apply variational inference only to the latter:
 
\begin{itemize}
    \item \textbf{Embedding network $g_\xi$ (frozen after pre-training):} The full U-Net \emph{except} its final layer, comprising the encoder, bottleneck, and decoder up to the last feature map.
    Given an input image $x \in \mathbb{R}^{C \times H \times W}$, it produces a dense embedding $g_\xi(x) \in \mathbb{R}^{D \times H \times W}$, preserving the full spatial resolution.
    This network is pre-trained to convergence on the segmentation task using a standard supervised loss, then frozen for all subsequent VIDS training.
    
    \item \textbf{Segmentation prediction head $f_\theta$ (stochastic):} A single $1\!\times\!1$ convolution with weight matrix $W \in \mathbb{R}^{C_\mathrm{out} \times D}$ and bias $b \in \mathbb{R}^{C_\mathrm{out}}$, such that $\theta = \mathrm{vec}(W, b)$ has dimension $d_\theta = D \cdot C_\mathrm{out} + C_\mathrm{out}$.
    For binary segmentation ($C_\mathrm{out} = 2$), this yields $d_\theta = 2D + 2$, which is small enough for amortized variational inference to be tractable.
    The prediction for image $x$ under parameters $\theta$ is
    \begin{equation}
        f_\theta\!\left(g_\xi(x)\right) = W \star g_\xi(x) + b \in \mathbb{R}^{C_\mathrm{out} \times H \times W},
        \label{eq:prediction_head}
    \end{equation}
    where $\star$ denotes a $1\!\times\!1$ convolution.
\end{itemize}
 
This decomposition means the inference network $h_\gamma$ only needs to produce parameters for the small $1\!\times\!1$ convolution rather than for the entire U-Net, making the approach computationally feasible. Figure~\ref{fig:architecture} summarizes this architecture
 
\paragraph{Spatial aggregation for the inference network}
In the original VIDS, embeddings are scalar vectors, and the training-set summary is simply the mean embedding $\bar{g}(x_{1:n}) = \frac{1}{n}\sum_{i=1}^{n} g_\xi(x_i)$.
For dense image embeddings $g_\xi(x_i) \in \mathbb{R}^{D \times H \times W}$, we first reduce each image to a global descriptor by spatial average pooling, yielding $\tilde{g}(x_i) \in \mathbb{R}^D$.
We then form a richer summary of the context set by concatenating the sample mean and sample standard deviation across images in the environment:
\begin{equation}
    \bar{g}(x_{1:n}) = \left[\frac{1}{n}\sum_{i=1}^{n} \tilde{g}(x_i)\ \Big\|\ \mathrm{std}_{i}\!\left(\tilde{g}(x_i)\right)\right] \in \mathbb{R}^{2D}.
    \label{eq:train_summary}
\end{equation}
Together with the test image embedding $g_\xi(x^*)$, these form the input to the inference network (Equation~\eqref{eq:inference_network}), which in VIDS-Seg takes dimension $4D$ input and outputs the mean $\mu \in \mathbb{R}^{d_\theta}$ and log-standard-deviation $\log\sigma \in \mathbb{R}^{d_\theta}$ of the variational posterior over the prediction-head parameters.
 
\paragraph{Segmentation-adapted energy function}
The original VIDS energy function integrates over the output space $\mathcal{Y}$.
For classification, this is a summation over class labels; for regression, a Monte Carlo integral over the real line.
For segmentation, the output is a spatial probability map, and the energy function is adapted accordingly.
For a predicted logit map $\hat{y} = f_\theta(g_\xi(x)) \in \mathbb{R}^{C \times H \times W}$, we compute the pixel-averaged log-probability under the predictive distribution:
\begin{equation}
    \mathcal{E}(\theta;\, x_{1:N}, x^*) = \sum_{i=1}^{N} \frac{1}{HW}\sum_{h,w} \log p(y_{hw} \mid x_i, \theta) + \frac{1}{HW}\sum_{h,w} \log p(y^*_{hw} \mid x^*, \theta).
    \label{eq:energy_segmentation}
\end{equation}
Averaging over spatial locations rather than summing ensures the energy scale is independent of image resolution, which is important for stability across datasets with different image sizes.
 
\paragraph{Segmentation log-likelihood}
The reconstruction term in the ELBO (Equation~\eqref{eq:elbo}) uses a pixel-wise cross-entropy loss, averaged over pixels and then summed over the batch:
\begin{equation}
    \log p(y_{1:N} \mid x_{1:N}, \theta) = -\frac{1}{N} \sum_{i=1}^{N} \sum_{h,w} \mathrm{CE}\!\left(f_\theta(g_\xi(x_i))_{hw},\, y_{i,hw}\right),
    \label{eq:log_lik_seg}
\end{equation}
where $y_{i,hw} \in \{0, 1\}$ is the ground-truth binary label at pixel $(h, w)$.
 
\subsection{Two-Stage Training Procedure}
\label{sec:training_procedure}
 
Training VIDS-Seg proceeds in two stages.
 
\paragraph{Stage 1: Pre-training the embedding network}
We train the full U-Net (embedding network $g_\xi$ plus a temporary $1\!\times\!1$ convolution head) on the segmentation task using a hybrid loss:
\begin{equation}
    \mathcal{L}_\mathrm{pre} = \alpha_\mathrm{CE}\, \mathcal{L}_\mathrm{CE} + (1 - \alpha_\mathrm{CE})\, \mathcal{L}_\mathrm{Dice},
    \label{eq:pretrain_loss}
\end{equation}
After convergence, the temporary head is discarded and the embedding network weights $\xi$ are frozen.
This stage establishes a strong, task-relevant feature representation before variational inference is introduced.
 
\paragraph{Stage 2: Variational fine-tuning with synthetic environments}
The inference network $h_\gamma$ and the stochastic prediction head parameters $\theta$ (sampled via the reparametrization trick) are jointly optimized using the cross-environment ELBO (Equation~\eqref{eq:cross_env_objective}).
The total loss is the negative mean ELBO plus a variance penalty (Equation~\eqref{eq:cross_env_objective}).
Only $h_\gamma$ is updated; $g_\xi$ remains frozen throughout the fine-tuning.
 
\subsection{Inference and Uncertainty Estimation}
\label{sec:inference_procedure}
At test time, given training images $\{x_i\}_{i=1}^{N}$ as the context set and a new image $x^*$, we estimate the segmentation and its associated uncertainty as summarized in Algorithm~\ref{alg:inference}.

\begin{algorithm}[t]
\caption{VIDS-Seg Inference and Uncertainty Estimation}
\label{alg:inference}
\begin{algorithmic}[1]
\Require Training images $\{x_i\}_{i=1}^{N}$, test image $x^*$, frozen embedding network $g_\xi$, inference network $h_\gamma$, number of samples $S$
\Ensure Predicted segmentation $\hat{y}$, entropy map $\mathcal{H}$
\State Compute context summary $\bar{g}(x_{1:N})$ via Equation~\eqref{eq:train_summary} \Comment{Context set embedding}
\State Compute test embedding $g_\xi(x^*)$
\State $(\mu, \log\sigma) \gets h_\gamma\big(\bar{g}(x_{1:N}), g_\xi(x^*)\big)$ \Comment{Variational posterior parameters}
\For{$s = 1$ \textbf{to} $S$}
    \State Sample $\theta^{(s)} \sim \mathcal{N}\big(\mu, \mathrm{diag}(\sigma^2)\big)$ \Comment{Reparametrization trick}
    \State $p^{(s)}_{:,h,w} \gets f_{\theta^{(s)}}\big(g_\xi(x^*)\big)_{:,h,w}$ \Comment{Per-sample probability map}
\EndFor
\State $\bar{p}_{hw,c} \gets \frac{1}{S}\sum_{s=1}^{S} p^{(s)}_{hw,c}$ \Comment{Mean predictive probability}
\State $\hat{y}_{hw} \gets \arg\max_c \bar{p}_{hw,c}$ \Comment{Final segmentation mask}
\State Compute entropy map $\mathcal{H} \in \mathbb{R}^{H \times W}$ from $\{p^{(s)}\}_{s=1}^{S}$ \Comment{Pixel-level uncertainty}
\State \Return $\hat{y}$, $\mathcal{H}$
\end{algorithmic}
\end{algorithm}

The entropy map $\mathcal{H}$ serves as our pixel-level uncertainty estimate, where high entropy indicates that the sampled parameter vectors disagree on the label at that location. By design of the adaptive prior, this should occur more frequently for OOD inputs.

\section{Experiments and Results}
\label{sec:experiments}
 
\subsection{Experimental Setup}
\label{sec:setup}
 
\subsubsection{Datasets}
\label{sec:datasets}

We evaluate our method on two publicly available datasets on echocardiography.

\paragraph{EchoNet-Dynamic \citep{ouyang2020video}}
EchoNet-Dynamic is a large-scale dataset of apical-four-chamber echocardiography videos acquired from adult patients at Stanford University Medical Center.
It comprises over 10,000 studies, each accompanied by frame-level expert tracings of the left ventricular (LV) endocardium at end-diastole (ED) and end-systole (ES), as well as ground-truth ejection fraction (EF) measurements.
We use this dataset as the \emph{in-distribution} (ID) training source, since all subjects are adults.
 
\paragraph{EchoNet-Pediatric \citep{reddy2023video}}
EchoNet-Pediatric contains echocardiography data from pediatric patients spanning a wide age range from infancy through adolescence.
Similar to EchoNet-Dynamic, the dataset provides LV tracings at ED and ES frames together with clinician-measured EF values.
We use the whole pre-processed data of the apical-4-chamber view of this dataset exclusively for evaluation, treating it as the target domain to study the model's behavior under covariate shift induced by the transition from adult to pediatric cardiac anatomy.
 
\subsubsection{Pre-processing and Segmentation Mask Generation}
\label{sec:preprocessing}
 
Neither dataset provides binary segmentation masks directly. Instead, annotations are given as contour-tracing representations of the LV boundary at the ED and ES frames. To obtain binary segmentation masks, we applied a contour-filling algorithm using the OpenCV library \citep{opencv_library}. We rasterized each tracing onto the corresponding video frame and filled the enclosed region to produce a pixel-wise binary mask.
 
To verify the geometric correctness of the resulting masks, we inverted the process by re-deriving the tracing representation from the binary mask and re-computing the EF estimate via Simpson's biplane method \citep{grossgasteiger2014image}. Subjects for whom the re-derived EF deviated by more than 5\% (absolute) from the reference EF provided in the dataset were excluded, ensuring that any segmentation artifacts introduced by the rasterization step did not corrupt the evaluation. Because tracings are only available at the ED and ES frames, segmentation masks, and therefore EF estimates, can be computed for exactly two frames per study.

\subsubsection{Baseline Methods and Training}
\label{sec:baselines_training}
 
All models are trained on EchoNet-Dynamic (adults only) and evaluated on EchoNet-Pediatric (children only) on a single NVIDIA A100 GPU.
 
\paragraph{Deep Ensemble}
We train an ensemble of $M=10$ U-Nets \citep{ronneberger_u-net_2015}, each with the same standard architecture but a different random initialization.
Each member is trained independently with the Adam optimizer (learning rate $1\!\times\!10^{-4}$) using a hybrid cross-entropy / soft-Dice loss ($\alpha_\mathrm{CE} = 0.5$) until convergence, with a maximum budget of 8 hours; all members converged within this limit.
At inference time, each member produces one probability map, and the $M$ maps are treated as equally-weighted samples for computing the predictive distribution and pixel-wise uncertainty.

\paragraph{PHiSeg \citep{shen_phiseg_2019}}
PHiSeg is a hierarchical probabilistic segmentation model that captures uncertainty through a multi-scale latent variable framework.
We use the official implementation with the default hyperparameters of \citep{shen_phiseg_2019}.
At inference time, 20 posterior samples are drawn to approximate the predictive distribution.

\paragraph{VIDS-Seg (Ours)}
Training proceeds in two stages.
\emph{Stage 1:} The U-Net backbone is pre-trained with the same hybrid loss and optimizer settings as the Ensemble members for 30 epochs.
\emph{Stage 2:} The backbone is frozen and the inference network $h_\gamma$ together with the $1\!\times\!1$ prediction head are fine-tuned for 30 epochs using the cross-environment ELBO (Equation~\eqref{eq:cross_env_objective}) with $L=40$ synthetic environments of size $n = 32$, KL weight $\lambda_\mathrm{KL} = 0.1$, and variance-penalty weight $\tau = 10^{-3}$. The framework is trained using the Adam optimizer with a learning rate of $1\!\times\!10^{-4}$. At inference time, $S = 20$ samples of $\theta$ are drawn from the variational posterior. Code will be made available upon acceptance.

\subsection{Results}
\label{sec:results}
 
\subsubsection{Identifying the OOD Subgroup: Segmentation Performance Across Age Groups}
\label{sec:ood_identification}
 
\paragraph{Motivation}
Children constitute a highly heterogeneous population: a teenager may be anatomically far more similar to an adult than an infant.
It is therefore not immediately obvious which pediatric subgroup, if any, should be considered out-of-distribution with respect to models trained on adults.
To identify the subgroup that poses the greatest challenge for adult-trained models, and thus constitutes our primary OOD evaluation population, we stratify EchoNet-Pediatric into five clinically meaningful age groups \citep{zubler2022evidence} and measure segmentation performance for each.

\paragraph{Evaluation protocol}
We define the following age strata: infants ($[0, 1)$~years, $n=304$), toddlers ($[1, 3)$~years, $n=442$), pre-schoolers ($[3, 6)$~years, $n=702$), school-age children ($[6, 13)$~years, $n=1982$), and teenagers ($[13, 18]$~years, $n=2366$), totaling $n=5796$ segmentations.
All models are trained solely on EchoNet-Dynamic and evaluated on each stratum.
Segmentation performance is measured by the Dice Similarity Coefficient (DSC) and the 95th percentile Hausdorff Distance (HD95) between the ground-truth binary mask and the model's main prediction, defined as the mean of the per-sample logit predictions thresholded at 0.5.
 
\begin{figure}[t]
    \centering
    \includegraphics[width=0.95\linewidth]{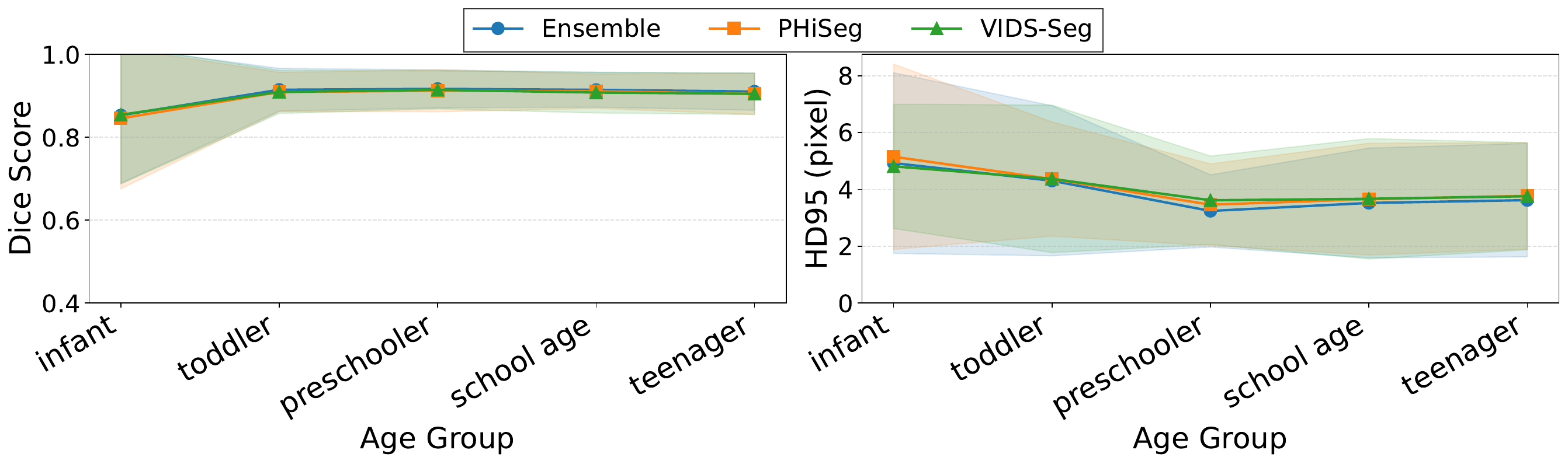}
    \caption{Mean $\pm$ standard deviation of Dice score (left) and 95th percentile Hausdorff Distance (right) for each pediatric age group. All models are trained on EchoNet-Dynamic (adults) and evaluated zero-shot on EchoNet-Pediatric.}
    \label{fig:age_dice}
\end{figure}

\paragraph{Findings}
Figure~\ref{fig:age_dice} reports DSC and HD95 for all methods across age groups.
All models achieve consistently high segmentation performance ($\mathrm{DSC} > 0.90$) on toddlers, pre-schoolers, school-age children, and teenagers.
In contrast, the infant group shows a pronounced and statistically significant performance drop to $\mathrm{DSC} \approx 0.84-0.85$ across all methods, along with markedly higher intra-group variance. Similarly, we observe an under-performance on the infant group for HD95 compared to the other groups. 
Crucially, this degradation is consistent across all three methods, indicating that the effect is driven by the data distribution rather than by any model-specific property. The consistent performance drop for infants supports the hypothesis that this subgroup is anatomically sufficiently distinct from adults to be considered OOD. Infants have proportionally different cardiac dimensions, higher heart rates, and distinct acoustic windows compared to older children or adults \citep{lopez2024guidelines}, all of which can plausibly cause a distribution mismatch with respect to features learned from EchoNet-Dynamic. We therefore designate the \emph{infant group as our OOD evaluation population} in all subsequent analyses.

\subsubsection{Uncertainty Quality for ID and OOD Populations}
\label{sec:uncertainty}

\paragraph{Motivation}
A practically useful uncertainty estimator should exhibit high uncertainty precisely where the model is likely to err, and this correspondence should be preserved both for in-distribution (ID) data and for OOD data.
Standard uncertainty methods are known to be overconfident in the presence of covariate shift. We investigate whether this is the case here and whether VIDS-Seg mitigates this problem.
A natural question is how VIDS-Seg's uncertainty estimates compare against post-hoc calibration baselines such as temperature scaling \citep{guo2017calibration}. Rather than treating temperature scaling as a competing UQ method in its own right, we use it to test a stronger hypothesis. If the baselines' poor spatial alignment is merely a symptom of poorly scaled confidence rather than a genuine failure to localize uncertainty, then calibrating them should close the gap to VIDS-Seg. We therefore additionally evaluate all three baselines after applying temperature scaling.

\paragraph{Evaluation protocol}
We evaluate uncertainty quality on two subgroups of EchoNet-Pediatric: the \emph{infant} group (OOD, $n=304$ segmentations) and \emph{non-infant} group (ID-proximal, $n=5492$ segmentations). As the primary metric we use the pixel-wise Normalized Cross-Correlation (NCC) between the per-pixel entropy map and a binary error map. The entropy at pixel $p$ is computed as $$\mathcal{H}_p = -\bar{s}_p \log \bar{s}_p - (1-\bar{s}_p)\log(1-\bar{s}_p),$$ where $\bar{s}_p$ is the mean predicted probability across $S$ samples ($S=20$ for PHiSeg and VIDS-Seg; $S=10$ for the Ensemble).
The error map is defined as the pixel-wise cross entropy between the mean logits and the ground-truth mask.
A higher NCC indicates better spatial alignment between uncertainty and prediction error which is the desired property for a trustworthy estimator.

To test whether the baselines' uncertainty maps merely require rescaling rather than being spatially uninformative, we further evaluate each baseline after \emph{temperature scaling} \citep{guo2017calibration} where a single scalar temperature $T$ is fitted per method on a held-out calibration split on the adult cohort by minimizing the negative log-likelihood of the predicted probabilities, and $\bar{s}_p$ is replaced by its temperature-rescaled counterpart before computing $\mathcal{H}_p$ and the resulting NCC. 

\paragraph{Findings}
Results are summarized in Table~\ref{tab:uncertainty}.
PHiSeg achieves the lowest NCC for both the non-infant and infant groups, suggesting that its uncertainty estimates are poorly calibrated spatially.
The Ensemble shows a clear improvement in NCC over PHiSeg for both populations.
VIDS-Seg attains the highest NCC among all methods for both groups, indicating superior spatial alignment between predicted uncertainty and actual prediction error.

Temperature scaling improves NCC for all three methods and in both subgroups, confirming that part of the baselines' raw NCC deficit is attributable to poorly scaled confidence rather than purely spatial misalignment. However, the ranking between methods is preserved after calibration: VIDS-Seg's calibrated NCC ($0.61 \pm 0.13$ non-infant, $0.62 \pm 0.11$ infant) remains higher than both temperature-scaled PHiSeg ($0.35 \pm 0.12$ non-infant, $0.33 \pm 0.12$ infant) and the temperature-scaled Ensemble ($0.50 \pm 0.15$ non-infant, $0.54 \pm 0.16$ infant). This indicates that VIDS-Seg's advantage is not an artifact of better-scaled confidence that a simple post-hoc rescaling could replicate, but reflects a genuinely better-localized correspondence between predicted uncertainty and prediction error.

The results suggest that common uncertainty quantification methods, even strong ones such as deep ensembles, tend to remain overconfident when confronted with OOD inputs, producing uncertainty maps that do not adequately reflect the elevated prediction error, and that this shortcoming persists even after standard post-hoc calibration.
VIDS-Seg, by contrast, is explicitly designed to model the uncertainty arising from distribution shift through its adaptive prior and inverse-bootstrapping training procedure, which appears to translate into a more spatially reliable uncertainty signal under distribution shift, even after standard post-hoc calibration is applied to all methods.

\begin{table}[h]
\centering
\caption{Spatial uncertainty quality (NCC $\uparrow$, mean $\pm$ std.\ across subjects) for non-infant (ID) and infant (OOD) subgroups, before (Raw) and after (Temp.-scaled) temperature scaling.}
\label{tab:uncertainty}
\begin{tabular}{lcccc}
\toprule
 & \multicolumn{2}{c}{Non-infant NCC $\uparrow$} & \multicolumn{2}{c}{Infant NCC $\uparrow$} \\
\cmidrule(lr){2-3} \cmidrule(lr){4-5}
Method & Raw & Temp.-scaled & Raw & Temp.-scaled \\
\midrule
Ensemble         & $0.35 \pm 0.15$ & $0.50 \pm 0.15$ & $0.39 \pm 0.17$ & $0.54 \pm 0.16$ \\
PHiSeg           & $0.14 \pm 0.11$ & $0.35 \pm 0.12$ & $0.11 \pm 0.09$ & $0.33 \pm 0.12$ \\
VIDS-Seg (Ours)  & $\mathbf{0.50 \pm 0.14}$ & $\mathbf{0.61 \pm 0.13}$ & $\mathbf{0.52 \pm 0.12}$ & $\mathbf{0.62 \pm 0.11}$ \\
\bottomrule
\end{tabular}
\end{table}

To assess whether the NCC improvement of VIDS-Seg over the Ensemble is statistically meaningful rather than an artifact of subgroup variance, we performed a Wilcoxon signed-rank test on paired per-subject NCC values, for both the raw (uncalibrated) predictions and the temperature-scaled predictions. For the uncalibrated predictions, the difference is significant for both the infant group ($p = 1.9 \times 10^{-8}$, rank-biserial effect size $r_{rb} = 0.86$) and the non-infant group ($p = 1.1 \times 10^{-170}$, $r_{rb} = 0.94$), indicating a large and consistent effect. For the calibrated predictions, the difference remains significant, with $p = 2.4 \times 10^{-5}$ ($r_{rb} = 0.68$) for the infant group and $p = 5.5 \times 10^{-135}$ ($r_{rb} = 0.83$) for the non-infant group. The effect sizes are somewhat smaller after calibration, consistent with temperature scaling closing part of the raw NCC gap. Nonetheless, the difference remains large and highly significant in both subgroups, confirming that VIDS-Seg's improvement over the Ensemble is not explained away by simple post-hoc recalibration.
We restrict formal significance testing to the VIDS-Seg vs.\ Ensemble comparison, as the Ensemble is the stronger of the two baselines in both raw and calibrated NCC. The corresponding gap to PHiSeg is larger throughout (Table~\ref{tab:uncertainty}) and we omit its statistics for brevity.
\subsubsection{Qualitative Analysis}
\label{sec:qualitative}
 
\begin{figure}[ht]
    \centering
    \includegraphics[width=0.9\linewidth]{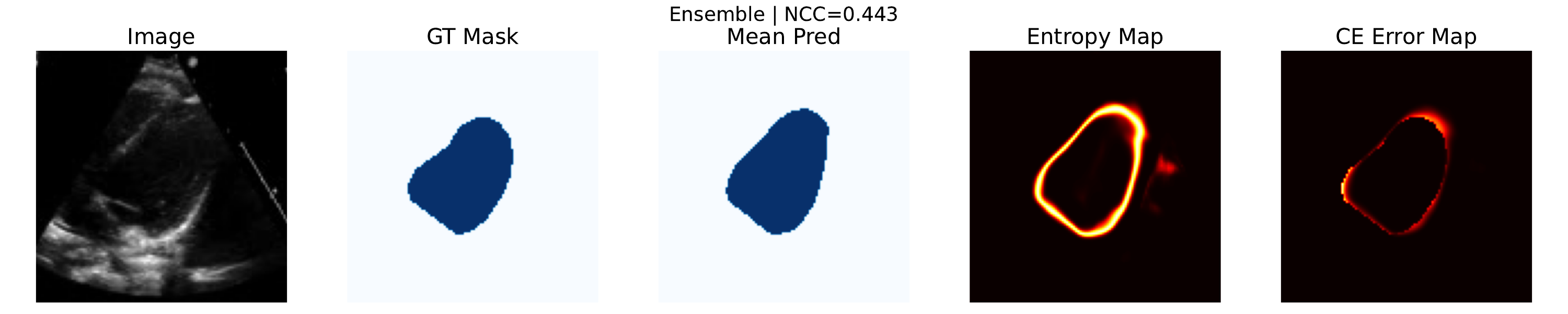}\\
    \includegraphics[width=0.9\linewidth]{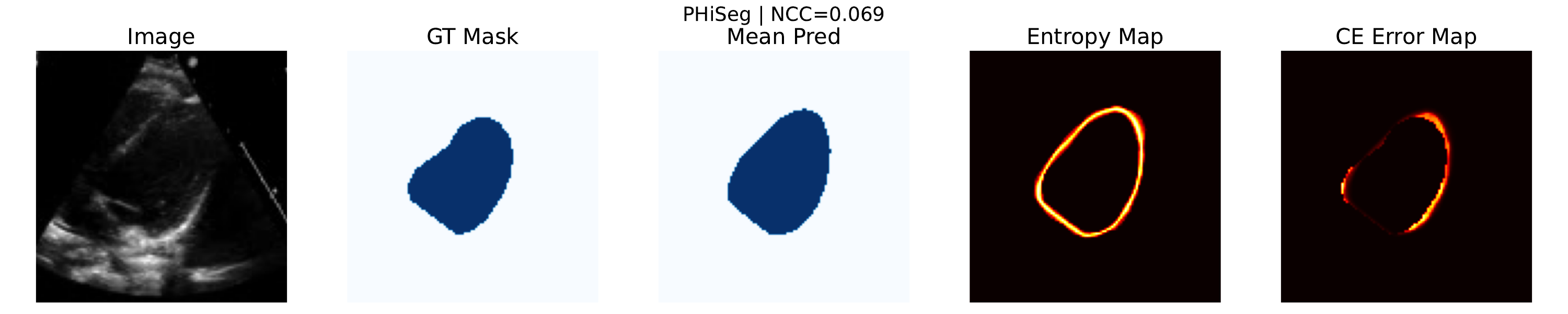}\\
    \includegraphics[width=0.9\linewidth]{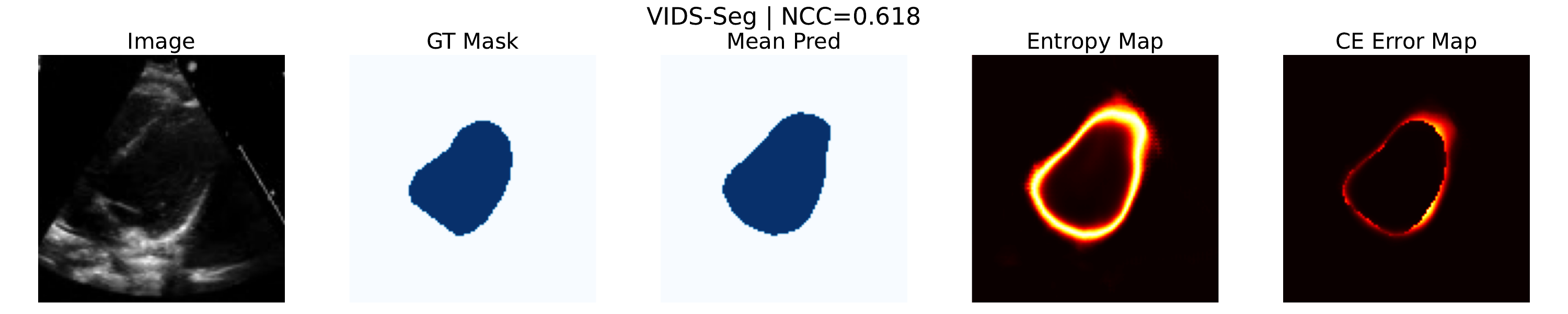}
    \caption{Qualitative comparison of uncertainty and error maps for a representative infant (OOD) case.Each row corresponds to one method: Deep Ensemble (NCC$=0.443$), PHiSeg (NCC$=0.069$), and VIDS-Seg (NCC$=0.618$). Columns show, from left to right: the input echocardiogram frame, the ground-truth left ventricular segmentation mask, the mean prediction, the pixel-wise entropy map, and the pixel-wise cross-entropy error map.}
    \label{fig:qualitative}
\end{figure}

Figure~\ref{fig:qualitative} shows representative examples of predictions, uncertainty (entropy) maps, and pixel-wise error maps for one infant (OOD) and one non-infant (ID) subject, for each method.
 
Across all three methods, the magnitude of the error map is broadly comparable, consistent with the quantitative findings in Section~\ref{sec:ood_identification} showing similar Dice scores for all models.
However, there are clear qualitative differences in how each method distributes its uncertainty.
 
For PHiSeg, the entropy map tends to be very narrow and only around the immediate border of the prediction.
The Ensemble produces more spatially focused uncertainty, with higher entropy at the LV boundary, but still misses many of the error-prone regions in the infant cases.
VIDS-Seg, in particular for the infant example, shows entropy that is more tightly concentrated at the location of prediction errors. This qualitative behavior directly mirrors the superior NCC scores reported in Table~\ref{tab:uncertainty}.

\subsubsection{Downstream Clinical Impact: Ejection Fraction Estimation and Cardiac Health Assessment}
\label{sec:ef}
 
\paragraph{Motivation}
In clinical practice, LV segmentations are rarely an end goal. They are typically used to derive quantitative biomarkers such as the ejection fraction, which directly informs cardiac health assessment and treatment decisions.
It is therefore important to assess not only segmentation quality but also the downstream impact of the model's predictions and their associated uncertainties.
 
\paragraph{EF estimation and cardiac health assessment protocol}
From each binary segmentation mask we recover a contour tracing and apply Simpson's biplane method to estimate the LV volume at ED and ES.
The EF is then computed as:
\begin{equation}
    \mathrm{EF} = \frac{V_{\mathrm{ED}} - V_{\mathrm{ES}}}{V_{\mathrm{ED}}} \times 100\%,
    \label{eq:ef}
\end{equation}
where $V_{\mathrm{ED}}$ and $V_{\mathrm{ES}}$ denote the end-diastolic and end-systolic volumes, respectively.
Beyond estimating EF as a continuous quantity, we also consider the clinically relevant task of identifying cardiac malfunction for the infant subgroup, which we define as $\mathrm{EF} < 50\%$. Framing this as a binary classification problem allows us to assess whether the predicted EF values are not only numerically accurate but also reliable for flagging patients who require closer clinical attention.
 
For the \textbf{Ensemble}, each of the 10 members produces one segmentation for each of the ED and ES frames, yielding 10 paired EF estimates; we report their mean.
For \textbf{PHiSeg} and \textbf{VIDS-Seg}, we draw $S=20$ samples from the posterior independently for ED and ES, since the samples across frames do not share a latent variable and therefore carry no natural pairing.
We form 20 EF estimates by randomly pairing ED and ES samples, then report the mean as the final EF prediction. Performance is evaluated by the Mean Absolute Error (MAE) between the predicted and ground-truth EF across each age stratum. 
 
\begin{figure}
    \centering
    \includegraphics[width=0.6\linewidth]{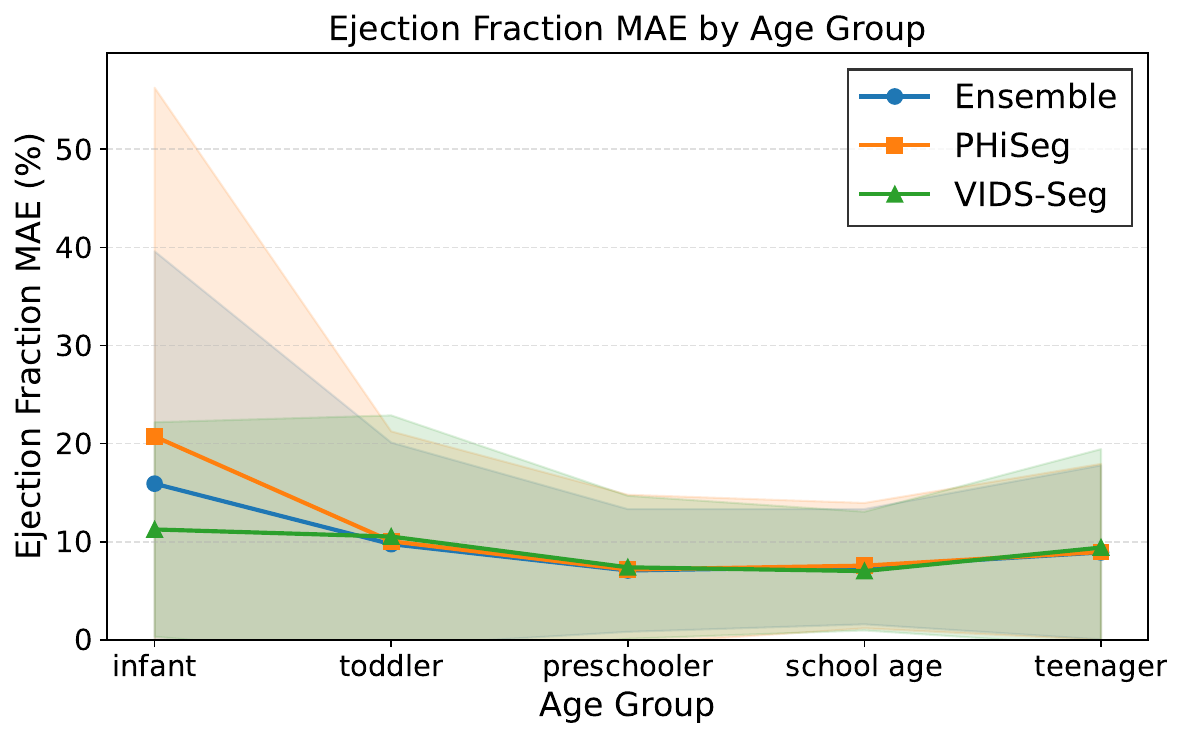}
    \caption{Mean Absolute Error (MAE, \%) $\pm$ standard deviation of ejection fraction estimates per age group. Lower is better.}
    \label{fig:ef_mae}
\end{figure}

\paragraph{Findings}
Results are shown in Figure~\ref{fig:ef_mae}.
For all non-infant age groups, the three methods perform comparably, exhibiting similar MAE values and standard deviations.
The infant group reveals a clear differentiation.
PHiSeg shows the highest MAE for infants and the largest variance across subjects, indicating that it not only makes larger errors but also does so inconsistently.
The Ensemble improves upon PHiSeg in mean MAE, but still exhibits substantial variance, suggesting that it frequently fails for individual infant cases.
VIDS-Seg achieves the lowest MAE for infants and, notably, substantially lower variance, demonstrating a more stable and reliable EF prediction even for this OOD population. In general, the MAE is a bit higher compared to other state-of-the-art methods for EF prediction \citep{akan2025viviechoformer}. We note that our model was not optimized for ejection fraction prediction directly which explains the elevated error compared to other methods.
This pattern is supported by the cardiac malfunction classification results. VIDS-Seg attains the highest AUROC (0.94), followed by the Ensemble (0.90), while PHiSeg trails behind with an AUROC of 0.81. The ranking mirrors the MAE findings in the infant subgroup: the higher variance and larger errors observed for PHiSeg translate into a reduced ability to correctly discriminate between healthy and impaired cardiac function, whereas VIDS-Seg's more stable EF estimates yield the most reliable classification of cardiac malfunction.

\section{Discussion and Conclusion}
\label{sec:discussion}
Systematic under-performance on OOD subgroups is a well-documented failure mode of clinical ML \citep{chatterjee2025children}, yet most UQ methods provide no mechanism to increase uncertainty precisely when the model is most likely to fail. We introduced VIDS-Seg, an extension of VIDS \citep{slavutsky2025quantifying} to dense segmentation, and demonstrated it on left ventricular segmentation in echocardiography across adult and pediatric cohorts.
Our three principal findings are: (1) infants are the only age group for which segmentation performance degrades consistently across all models, justifying their treatment as OOD; (2) VIDS-Seg produces significantly higher spatial correspondence between uncertainty and prediction error on this OOD group compared to deep ensembles and PHiSeg, \emph{without} sacrificing segmentation performance; and (3) this uncertainty quality improvement translates directly into more stable ejection fraction estimates for infants, demonstrating a positive effect on downstream clinical biomarkers.

The concentration of under-performance in infants, rather than in children broadly, is clinically meaningful rather than incidental. Cardiac dimensions and left ventricular geometry scale non-linearly with age, and infants additionally present with higher heart rates and distinct acoustic windows compared to older children or adults \citep{lopez2024guidelines}, all of which plausibly place this subgroup anatomically furthest from the adult training distribution. This pattern is consistent with reports of analogous pediatric performance gaps in other clinical imaging tasks, including organ segmentation and chest radiograph interpretation \citep{chatterjee2025children, shin2022diagnostic}, suggesting that infant-specific distribution shift is a recurring rather than dataset-specific challenge for models trained on adult populations.

The uncertainty quality ordering, PHiSeg $<$ Ensemble $<$ VIDS-Seg, holds across both ID and OOD populations and has a natural interpretation. PHiSeg was designed to capture aleatoric uncertainty from annotation ambiguity rather than covariate shift. Deep ensembles implicitly capture epistemic uncertainty through initialization diversity, but without an explicit OOD mechanism. VIDS-Seg is the only method with a principled posterior that widens in response to distributional novelty. Notably, this ordering is not simply an artifact of the baselines' confidence being poorly scaled: when we apply post-hoc temperature scaling to all three methods, NCC improves across the board, yet VIDS-Seg's advantage over the calibrated Ensemble and calibrated PHiSeg remains large and statistically significant (Section~\ref{sec:uncertainty}). This indicates that VIDS-Seg's uncertainty maps are better spatially localized, not merely better scaled in magnitude — a distinction that a simple recalibration step, such as one might apply as a lightweight fix to an existing deployed model, cannot replicate. While VIDS-Seg's uncertainty-error correspondence is the strongest, it may not yet suffice as a standalone alert signal without further calibration. Post-hoc frameworks such as conformal prediction \citep{angelopoulos_gentle_2022} or RCPS \citep{bates_distribution-free_2021} could be applied on top, though their global coverage guarantees explicitly do not hold on OOD data, due to their assumption of exchangeability.

The downstream clinical relevance of this uncertainty quality gap is most apparent in the ejection fraction results. VIDS-Seg not only achieves the lowest EF estimation error for infants but also the highest AUROC (0.94) for detecting cardiac malfunction ($\mathrm{EF} < 50\%$), compared to 0.90 for the Ensemble and 0.81 for PHiSeg. This is a clinically more legible result than EF MAE alone. A clinician ultimately cares less about the exact percentage-point error of an EF estimate than about whether a patient requiring closer follow-up is correctly flagged. The consistency between the uncertainty quality ranking, the EF error ranking, and the malfunction classification ranking suggests that better-localized uncertainty is not merely a diagnostic property of the model but one that propagates through the full pipeline to the metric that actually matters at the point of care.

\paragraph{Clinical utility}
The most direct clinical implication of this work is that VIDS-Seg can serve as an \emph{uncertainty-based safety layer} on top of any deployed segmentation model. When the predicted entropy exceeds a threshold, the clinician is alerted that the case may lie outside the model's reliable operating range, prompting expert review before downstream biomarkers are acted upon. It requires no retraining, no additional labeled data, and no modifications to the core segmentation network. A key remaining step is translating pixel-level entropy into a clinically interpretable scalar, such as a predicted confidence interval on ejection fraction, to make the uncertainty signal actionable without requiring clinicians to interpret spatial maps. Prospective evaluation with clinical partners is the ultimate test of whether this alert mechanism delivers on its promise of safer, more robust ML deployment.

VIDS-Seg takes a step towards clinical ML that can communicate its own limitations. Not by eliminating distributional gaps, but by making them visible. Future work should explore uncertainty propagation along the full diagnostic pipeline, extension to multi-organ settings, and co-design of uncertainty-driven workflow integrations with clinical stakeholders.

\section*{CRediT Authorship Contribution Statement}
\textbf{Paul Fischer:} Writing – review \& editing, Writing – original draft, Visualization, Validation, Software, Methodology, Investigation, Formal analysis, Conceptualization. \textbf{Ece Özkan Elsen:} Supervision, Writing – review \& editing, Validation, Methodology, Conceptualization, Resources, Project administration. 

\section*{Declaration of Competing Interest}
The authors declare that they have no known competing financial interests or personal relationships that could have appeared to influence the work reported in this paper.

\section*{Acknowledgments}
The authors thank Yuli Slavutsky for helpful discussions on the original VIDS method and for providing code and optimization advice.








\bibliographystyle{elsarticle-num}  
\bibliography{bibliography}
\end{document}